\documentclass{article}

\usepackage[preprint]{neurips_2026}

\usepackage[utf8]{inputenc}
\usepackage[T1]{fontenc}
\usepackage{amsmath}
\usepackage{amssymb}
\usepackage{amsfonts}
\usepackage{amsthm}
\usepackage{mathtools}
\usepackage{graphicx}
\usepackage{subcaption}
\usepackage{booktabs}
\usepackage{multirow}
\usepackage{makecell}
\usepackage{array}
\usepackage{url}
\makeatletter\g@addto@macro\UrlBreaks{\do\-}\makeatother
\usepackage{siunitx}
\usepackage{algorithm}
\usepackage{algpseudocode}
\usepackage[table]{xcolor}
\usepackage{microtype}

\usepackage{tikz}
\usetikzlibrary{arrows.meta,positioning,fit,calc}

\theoremstyle{definition}
\newtheorem{remark}{Remark}

\newif\ifshowedits
\showeditsfalse

\usepackage[hidelinks]{hyperref}

\DeclareMathOperator*{\argmin}{arg\,min}

\newlength{\narrowfigw}
\makeatletter
\AtBeginDocument{%
  \if@twocolumn\setlength{\narrowfigw}{\columnwidth}%
  \else\setlength{\narrowfigw}{0.72\textwidth}\fi}
\makeatother

\title{Imitation Learning of Avoidance Modes for Efficient \\ Multimodal Model Predictive Control}

\author{%
  Changyu Lee\thanks{Corresponding author.} \\
  Kongju National University \\
  \texttt{leeck@kongju.ac.kr} \\
  \And
  Byung-Kwan Lee \\
  NVIDIA \\
  \texttt{byungkwanl@nvidia.com} \\
  \And
  Kanghoon Yoon \\
  KAIST \\
  \texttt{ykhoon08@kaist.ac.kr} \\
}

\begin{document}

\maketitle

\begin{abstract}
Multimodal model predictive control (MM-MPC) avoids local minima among moving obstacles by solving one optimal control problem (OCP) per combination of avoidance sides, so its computational cost grows exponentially with the number of decision-critical obstacles. We propose SideNet-MPC, which replaces this enumeration with a mode decision learned by imitating MM-MPC as an expert. A graph neural network is trained to predict the expert's minimum-cost mode, one avoidance side per obstacle. Its edges encode, for each obstacle pair, the gap predicted for the moment the vehicle first reaches either obstacle, so a neighbor that blocks a preferred passage can reverse the proposed side. The expert's selection rule compares its minimum-cost mode with the previously executed mode; SideNet-MPC keeps this rule and substitutes the network proposal for the minimum-cost mode. Hence at most two OCPs are solved per control period, and every applied input still comes from an OCP with control barrier function constraints rather than from the network. In $100$ seed-matched closed-loop Monte Carlo runs of mixed traffic, SideNet-MPC reaches $93\%$ of the expert's success rate and needs \SI{18.6}{ms} per control period on a single thread, 48\% less than the expert's \SI{35.5}{ms} on sixteen parallel threads. In a controlled gap sweep, its proposed side reverses in the same gap interval as the expert's, whereas proposers without pairwise edges, trained on the same data, never reverse and all deadlock at the blocked gates of a deterministic scene.
The project page is available at \href{https://sail-knu.github.io/Imitation-Learning-of-Avoidance-Modes-for-Computationally-Efficient-MM-MPC/}{\textcolor{blue}{link}}.
\end{abstract}

\section{Introduction}
\label{sec:intro}

 A standard navigation architecture decomposes the problem into a global planner that provides a reference path and a local planner that reacts to unmapped or dynamic obstacles with methods such as the dynamic window approach \citep{fox2002dynamic}, artificial potential fields \citep{khatib1986real,zhang2023obstacle}, or model predictive control (MPC) with distance-based safety constraints \citep{wang2018design,brito2019model,mpc9_lee2023nonlinear}. When the reference path is not collision-free, for instance a straight centerline through moving traffic, the local planner itself must decide the avoidance topology, that is, whether to pass each obstacle on the left or on the right. Trajectories in different homotopy classes cannot be deformed into one another without crossing an obstacle \citep{bhattacharya2010search}. Gradient-based optimization therefore typically remains in the class of its initialization, so standard MPC is prone to local minima and deadlock when several obstacles constrain the same passage.

\begin{figure}[!t]
\centering
\begin{subfigure}{\narrowfigw}
\centering
\includegraphics[width=0.85\linewidth]{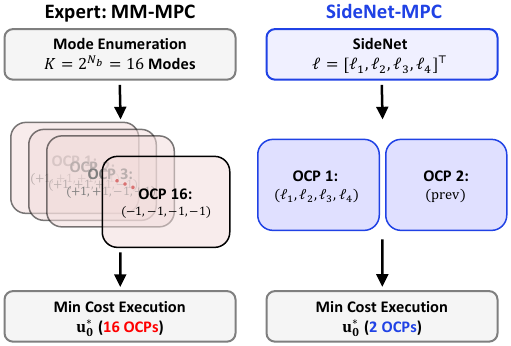}
\caption{Block diagrams of the expert and SideNet-MPC.}
\end{subfigure}
\vspace{0.25em}
\begin{subfigure}{\narrowfigw}
\centering
\includegraphics[width=\linewidth]{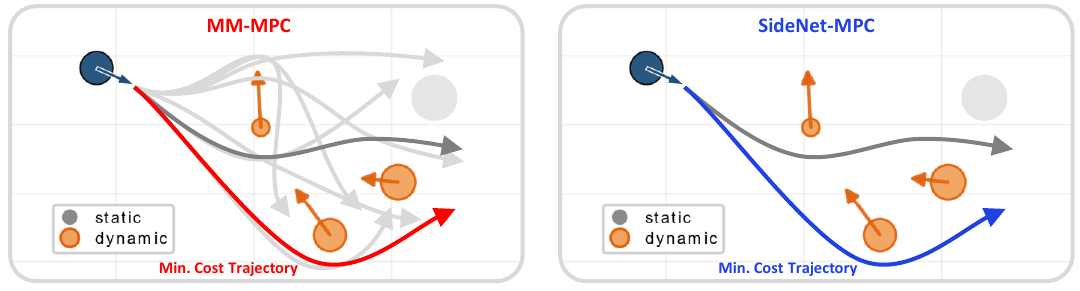}
\caption{Candidate (gray) and executed trajectories of the expert (red) and SideNet-MPC (blue) at one planning step of the same scene.}
\end{subfigure}
\caption{Overview of SideNet-MPC. Both controllers apply the same selection
rule to the previously executed mode and one challenger, and differ only in how
the challenger is produced: the MM-MPC expert solves all $K=2^{N_b}=16$ OCPs and takes the minimum-cost mode, whereas SideNet-MPC solves only the mode proposed by the network, so at most two OCPs are solved per control
period.}
\label{fig:overview}
\end{figure}

To escape these local minima, one line of work maintains admissible candidate trajectories of distinct topologies and optimizes them in parallel \citep{kuderer2014online,rosmann2017integrated,zheng2025barrier}; another reintroduces a high-level planner that generates homotopy guide paths for the optimizer \citep{de2024topology,li2025trust}. A third line of work instead fixes the topology through the constraints. The turning circle-based control barrier function (TC-CBF) \citep{arxiv-tccbf,lee2025efficient} encodes the avoidance side directly in the barrier: keeping the left or right turning circle clear of an obstacle commits the vehicle to passing on that side while respecting its nonholonomic turning limits. Building on this, multimodal MPC (MM-MPC) \citep{lee2026multimodal} enumerates all $K=2^{N_b}$ combinations of avoidance sides, called modes, for the $N_b$ decision-critical obstacles, called branch obstacles. It solves one TC-CBF constrained optimal control problem (OCP) per mode and selects the executed mode from the optimal costs. The topology is thus chosen by an explicit cost comparison, which reduces reliance on the optimizer's initialization and requires no homotopy guide path.

The difficulty is computational. Each control period requires $K$ OCP solves, and $K=2^{N_b}$ grows exponentially with the number of branch obstacles. The same scaling appears in parallel homotopic planners \citep{zheng2025barrier} and in contingency and scenario-tree MPC, whose trajectory trees grow with the number of futures they hedge against \citep{alsterda2019contingency,chen2022interactive}. The OCPs are independent and can be solved in parallel, but parallelization reduces only the wall-clock time, not the total computation. In our setting, $N_b=4$ yields $K=16$ OCPs, and solving them fully in parallel needs sixteen CPU threads, which a workstation provides but typical embedded computers do not. This paper therefore asks whether the enumeration can be replaced by a mode decision learned by imitating MM-MPC as an expert, without giving up the expert's structural property: the executed input always comes from an OCP carrying TC-CBF constraints.

Learning has been widely used to reduce the online cost of optimization-based control \citep{hewing2020learning}. Approximate MPC directly regresses the control law \citep{hertneck2018learning,karg2020efficient,liu2026approximate}, which removes online optimization together with its explicit constraint handling. Learned policies are therefore often paired with a predictive safety filter that minimally modifies their inputs to keep the constraints satisfied \citep{wabersich2021predictive}. Learned warm-starting instead retains online optimization, predicting active sets \citep{klauco2019machine} or approximating optimal trajectories to initialize a nonlinear predictive controller \citep{mansard2018using}. When the problem has several homotopy classes, however, a warm start also decides the class implicitly, without the cost comparison that the expert performs. This motivates learning the class decision explicitly rather than leaving it to the initialization.

Closest to this work are methods that learn the discrete decisions of an optimization or planning problem. \citet{bertsimas2022online} predict the discrete strategy of a mixed-integer program so that only a convex problem remains to be solved online. CoCo \citep{cauligi2022coco} learns such strategies for combinatorial motion planning, but its obstacle-avoidance demonstration uses static axis-aligned obstacles and does not examine mode selection among moving obstacles. \citet{gasse2019exact} use a graph convolutional network to imitate strong branching, an expensive expert rule for variable selection in branch-and-bound. \citet{reiter2025equivariant} address dynamic traffic with an ensemble of recurrent permutation-equivariant deep sets: each network proposes a candidate that is evaluated through a soft-constrained quadratic program, and the lowest-cost trajectory is then projected onto the feasible set.

We propose SideNet, a learned mode proposer, and SideNet-MPC, the controller built on it; Fig.~\ref{fig:overview} contrasts SideNet-MPC with the enumerating expert. SideNet is trained to predict the expert's minimum-cost mode, one left/right decision per branch obstacle. The expert's selection rule compares the minimum-cost mode with the previously executed mode; SideNet-MPC keeps this rule and substitutes the network proposal for the minimum-cost mode, the only term that requires enumeration. A learned decision thus reaches the vehicle only through the expert's nonlinear TC-CBF OCP and selection rule. This matters because a wrong avoidance side is executed by the vehicle, whereas a poor imitated decision in the branch-and-bound setting of \citet{gasse2019exact} costs only solver time. In place of the ensemble of \citet{reiter2025equivariant}, SideNet-MPC uses one network and the previously executed mode, and applies the solution selected by the expert's rule without a separate projection step.

\begin{figure}[t!]
\centering
\begin{subfigure}{0.49\linewidth}
\includegraphics[width=\linewidth]{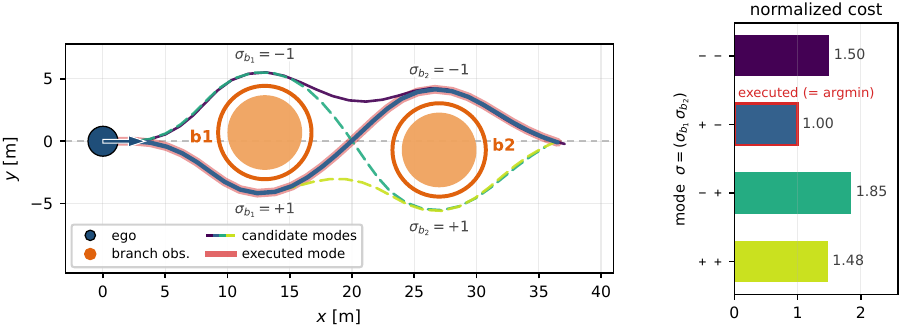}
\caption{Two branch obstacles on the centerline, $K=2^2$ modes.}
\label{fig:modes-a}
\end{subfigure}
\begin{subfigure}{0.49\linewidth}
\includegraphics[width=\linewidth]{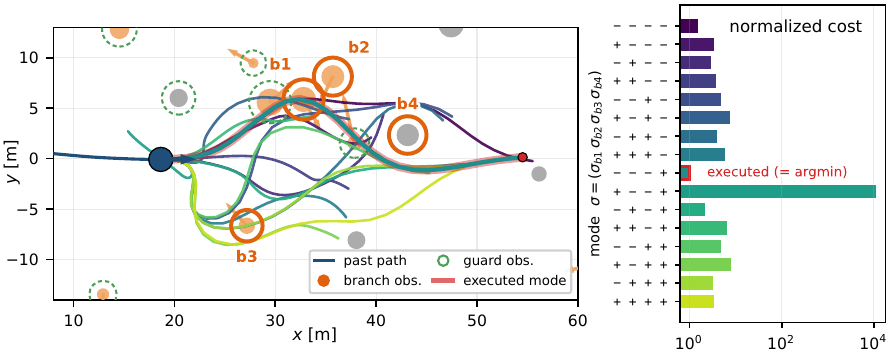}
\caption{Cluttered scene with four branch obstacles, $K=2^4$ modes.}
\label{fig:modes-b}
\end{subfigure}
\caption{Avoidance modes enumerated by the expert at one planning step. Each
mode fixes the left/right side of every branch obstacle and yields
one TC-CBF constrained OCP; the minimum-cost mode is shown in red. In (a)
the four local optima lie in different homotopy classes.}
\label{fig:modes}
\end{figure}

The side decision also depends on the obstacles around each branch obstacle: a neighbor can close the passage that the isolated choice would use. SideNet is therefore a permutation-equivariant graph neural network (GNN) whose edge features include the surface gap of each obstacle pair, predicted for the moment the vehicle first reaches either obstacle of the pair. \citet{reiter2025equivariant} pool obstacle features in equivariant deep-set layers \citep{zaheer2017deep}; SideNet instead uses edge-conditioned message passing \citep{gilmer2017neural}, so the gap of each pair enters the message explicitly rather than being inferred from pooled obstacle features.

The main contributions of SideNet-MPC are summarized as follows:
\begin{itemize}
\item A learned mode decision that replaces only the enumeration of the MM-MPC expert: a network proposal substitutes for the minimum-cost mode in the expert's selection rule, so at most two TC-CBF OCPs are solved per control period for any number of branch obstacles, and every applied input still comes from the expert's OCP.
\item A permutation-equivariant GNN whose edge features encode the surface gap of each obstacle pair predicted for the moment the vehicle first reaches either obstacle. In a controlled gap sweep, its proposed side reverses in the same gap interval as the expert's, whereas proposers without pairwise edges trained on the same data, including one with a matched parameter count, keep one side regardless of the gap.
\item A closed-loop Monte Carlo evaluation on seed-matched mixed traffic, in which SideNet-MPC retains $93\%$ of the expert's success rate and reduces the computation time per control period from \SI{35.5}{ms} on sixteen parallel threads to \SI{18.6}{ms} on a single thread.
\end{itemize}

The remainder of this paper is organized as follows. Section~\ref{sec:prelim} reviews the TC-CBF and the MM-MPC expert and states the problem. Section~\ref{sec:method} introduces SideNet and its training. Section~\ref{sec:deploy} presents SideNet-MPC, which places the SideNet proposal and the previously executed mode as two candidates in the expert's selection rule. Section~\ref{sec:exp} reports the simulation studies, and Section~\ref{sec:conclusion} concludes the paper.

\section{Preliminaries and Problem Statement}
\label{sec:prelim}

\subsection{Vehicle Model}

We consider a five-state unicycle model with state $\mathbf{x}=[x,y,\psi,u,r]^\top$ and input $\mathbf{u}=[a,\dot r]^\top$,
\begin{equation}
\dot{\mathbf{x}} = [\,u\cos\psi,\; u\sin\psi,\; r,\; a,\; \dot r\,]^\top,
\label{eq:model}
\end{equation}
where $(x,y)$ is the position, $\psi$ the heading, $u$ the speed, $r$ the yaw rate, $a$ the acceleration, and $\dot r$ the yaw acceleration.

\subsection{Turning Circle-based CBF}
\label{sec:tccbf}

A control barrier function (CBF) $h$ certifies forward invariance of the safe set $\{\mathbf{x}: h(\mathbf{x})\ge 0\}$ \citep{ames2016control,ames2019control}. In discrete time the condition $h(\mathbf{x}_{k+1})\ge(1-\gamma)h(\mathbf{x}_k)$ with $\gamma\in(0,1]$ can be imposed as an MPC constraint \citep{zeng2021safety}. Distance-based barriers, which keep the vehicle--obstacle distance above a required separation, are the common choice \citep{jian2023dynamic,jin2022collision} but encode no avoidance direction.

The TC-CBF \citep{arxiv-tccbf} instead measures the clearance of the vehicle's turning circles. Write $\mathbf{p}=[x,y]^\top$ for the vehicle position. For avoidance side $\sigma\in\{-1,+1\}$, the turning circle center is displaced laterally from the vehicle by the protected turning radius $R=c_R\,u/r_{\max}$, where $c_R$ inflates the turning radius for
margin,
\begin{equation}
\mathbf{c}(\mathbf{x},\sigma)=\mathbf{p}+R\big[\cos(\psi-\tfrac{\pi}{2}\sigma),\;
\sin(\psi-\tfrac{\pi}{2}\sigma)\big]^\top,
\end{equation}
so that the offset $\psi-\tfrac{\pi}{2}\sigma$ places the center on the right for $\sigma=+1$ and on the left for $\sigma=-1$. The barrier for obstacle $j$ with radius $\rho_j$ and position $\mathbf{o}_j$ is
\begin{equation}
h_j(\mathbf{x},\sigma;\mathbf{o}_j)=\big\lVert \mathbf{c}(\mathbf{x},\sigma)-\mathbf{o}_j
\big\rVert-(R_s+\rho_j+\delta+R),
\label{eq:tccbf}
\end{equation}
with vehicle radius $R_s$ and bumper margin $\delta$. Throughout, $\sigma=+1$
denotes passing the obstacle on the right: the right turning circle is kept
clear, so the obstacle stays on the vehicle's left side
(Fig.~\ref{fig:modes-a}). 

\subsection{Multimodal MPC Expert}
\label{sec:mmmpc}

The MM-MPC framework of \citet{lee2026multimodal} serves as the expert. Let $T_s$ be the control period,
$N$ the horizon length and $T_h=N T_s$, and let obstacle $j$ move with velocity $\mathbf{v}_j$. At each control period, the expert selects a branch set $\mathcal{B}$ of up to $N_b$ obstacles, whose avoidance sides are enumerated, and a guard set $\mathcal{G}$ of up to $N_g$ obstacles, whose sides are fixed by the relative bearing so that they constrain every mode without multiplying the number of modes.

The two sets are selected by a two-zone rule. For obstacle $j$, write
$\mathbf{p}_{\mathrm{rel},j}=\mathbf{o}_j-\mathbf{p}$ for the relative position,
$d_j=\lVert\mathbf{p}_{\mathrm{rel},j}\rVert-\rho_j$ for the surface distance,
and
$\beta_j=\mathrm{wrap}\bigl(\mathrm{atan2}(o_{j,y}-y,\,o_{j,x}-x)-\psi\bigr)$
for the relative bearing. Obstacles within the sensing range $\lVert\mathbf{p}_{\mathrm{rel},j}\rVert\le d_{\mathrm{sens}}$ form the decision zone if they lie in the decision cone $|\beta_j|\le\theta_d/2$ and the peripheral zone otherwise. Every
decision-zone obstacle receives the urgency score
\begin{equation}
S_j = \frac{w_d}{d_j} + w_{\psi}\cos\beta_j,
\label{eq:select}
\end{equation}
which is large when the obstacle is both close and ahead. The decision-zone
obstacles with the $N_b$ largest $S_j$ fill the branch slots of $\mathcal{B}$. The $N_g$ guard slots of $\mathcal{G}$ are filled first from the remaining decision-zone obstacles and then from the peripheral zone, each in ascending surface distance. A guard obstacle keeps the side of its bearing: $\sigma_j=+1$ if $\beta_j>0$ and $\sigma_j=-1$ otherwise. Each slot carries an active flag $a_j\in\{0,1\}$, equal
to one when the slot is filled; for an unfilled slot ($a_j=0$) the barrier
constraint is made vacuous in the OCP and the features are zeroed at the
network input.

A mode is a sign vector $m=(\sigma_1,\ldots,\sigma_{N_b})\in\{-1,+1\}^{N_b}$, one entry per branch slot, so there are $K=2^{N_b}$ modes: four for $N_b=2$ in Fig.~\ref{fig:modes-a} and sixteen for $N_b=4$ in Fig.~\ref{fig:modes-b}. Write $\mathbf{e}_i=\mathbf{x}_i-\mathbf{x}_i^{\mathrm{ref}}$ for the tracking error, with $\lVert\mathbf{e}\rVert_Q^2=\mathbf{e}^{\top}Q\mathbf{e}$. The reference is the constant-speed centerline profile $\mathbf{x}_i^{\mathrm{ref}}=[\min(x+u_{\mathrm{des}}T_s i,\;x_{\mathrm{goal}}),\;0,\;0,\;u_{\mathrm{des}},\;0]^{\top}$, $i=0,\ldots,N$, where $x$ is the current longitudinal position, and is rebuilt at every control period. Dynamic obstacles are predicted at stage $i$ by $\mathbf{o}_{j,i}=\mathbf{o}_{j,0}+\mathbf{v}_j\,iT_s$. Slack is collected in $\mathbf{s}_i=[s_{1,i},\ldots,s_{N_b+N_g,i}]^{\top}$. For each mode the expert solves, for $i=0,\ldots,N-1$ and $j\in\mathcal{B}\cup\mathcal{G}$,
\begin{subequations}
\label{eq:ocp}
\begin{align}
J^*(m)=\min_{\mathbf{x},\mathbf{u},\mathbf{s}}\;&
\sum_{i=0}^{N-1}L(\mathbf{x}_i,\mathbf{u}_i,\mathbf{s}_i)
+\lVert\mathbf{e}_N\rVert_Q^2\\
\text{s.t.}\;\;
&\mathbf{x}_0=\mathbf{x}_{\mathrm{init}},\quad
\mathbf{x}_{i+1}=f_d(\mathbf{x}_i,\mathbf{u}_i),\\
&\mathbf{x}_i\in\mathcal{X},\quad \mathbf{u}_i\in\mathcal{U},\quad \mathbf{s}_i\ge 0,\\
&h_j(\mathbf{x}_{i+1},\sigma_j;\mathbf{o}_{j,i+1})
\ge(1-\gamma)\,h_j(\mathbf{x}_i,\sigma_j;\mathbf{o}_{j,i})
- s_{j,i},
\label{eq:ocp_cbf}
\end{align}
\end{subequations}
where
$L(\mathbf{x}_i,\mathbf{u}_i,\mathbf{s}_i)
=\lVert\mathbf{e}_i\rVert_Q^2
+\lVert\mathbf{u}_i\rVert_{R_u}^2
+\phi(\mathbf{s}_i)$
is the stage cost and $f_d$ is a fourth-order Runge--Kutta discretization of~\eqref{eq:model}. The state set $\mathcal{X}$ bounds the lateral position by the corridor walls, the speed by $u\in[0,u_{\max}]$ and the yaw rate by $|r|\le r_{\max}$; the input set $\mathcal{U}$ bounds $|a|\le a_{\max}$ and $|\dot r|\le\dot r_{\max}$. In~\eqref{eq:ocp_cbf}, $\sigma_j$ is given by the mode $m$ for branch obstacles and by the bearing rule above for guard obstacles. The slack lets modes that would otherwise violate the CBF constraints remain solvable, and its penalty $\phi(\mathbf{s})=\lambda_q\lVert\mathbf{s}\rVert_2^2+\lambda_l\lVert\mathbf{s}\rVert_1$ gives such modes a large cost. 
The $K$ OCPs are independent and are solved in parallel with acados \citep{verschueren2022acados} and CasADi \citep{casadi}. A solve is accepted if the sequential quadratic programming (SQP) solver reports success or if both the equality and inequality constraint residuals are below an absolute tolerance of $5\times10^{-2}$; stationarity is not checked, so an accepted iterate may be feasible but suboptimal. A rejected mode is assigned infinite cost.

The expert's selection rule chooses the executed mode from two candidates: the minimum-cost mode
$m^{\mathrm{best}}=\argmin_m J^*(m)$ and the previously executed mode
$m^{\mathrm{prev}}$, re-associated with the current branch obstacles
(Section~\ref{sec:deploy}). To avoid
frequent mode switching, the previous mode is retained when its cost is
sufficiently close to that of the minimum-cost mode:
\begin{equation}
\begin{split}
\text{keep } m^{\mathrm{prev}} \iff{}&
J^*(m^{\mathrm{prev}})\le J^*(m^{\mathrm{best}})\\
&+\eta\,\max\!\Big(\tfrac{J^*(m^{\mathrm{prev}})+J^*(m^{\mathrm{best}})}{2},\,1\Big).
\end{split}
\label{eq:hyst}
\end{equation}
Here $\eta=0.2$ is a relative tolerance on the mean of the two costs, floored
at one cost unit so that the tolerance does not vanish when both costs are
small.

\subsection{Problem Statement}
\label{sec:problem}

The expert of Section~\ref{sec:mmmpc} solves $K=2^{N_b}$ OCPs per control period. The
goal of this paper is a controller that (i) solves a bounded number of OCPs
per control period, independent of $N_b$; (ii) preserves the structural property that
executed inputs are produced by OCPs carrying the TC-CBF constraints of the
selected obstacles, rather than by a learned function; and (iii) retains as
much of the expert's closed-loop success as possible in encounters where the
avoidance side matters.

\begin{figure}[t]
\centering
\includegraphics[width=\linewidth]{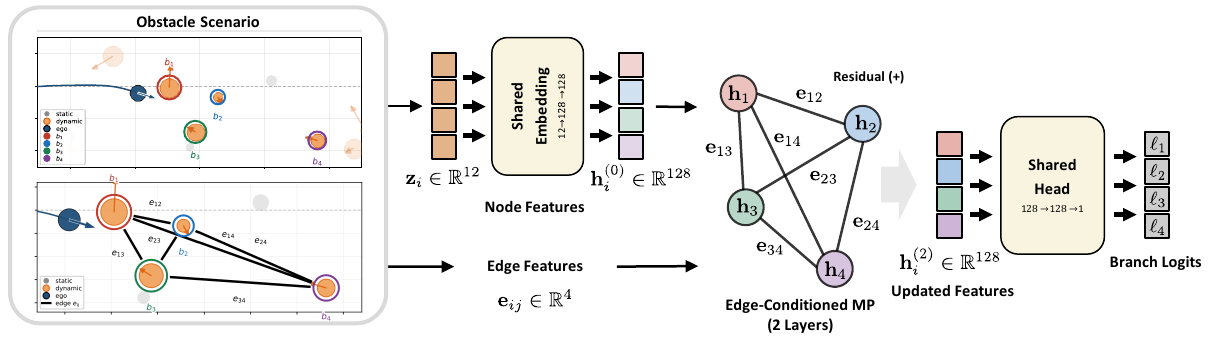}
\caption{SideNet, the mode proposer. The $N_b$ branch slots form a complete
graph. A shared embedding maps each slot input $\mathbf{z}_i$~\eqref{eq:node}
to $\mathbf{h}_i$, and each pair carries the edge features
$\mathbf{e}_{ij}$~\eqref{eq:edge}, including the gap $g^*_{ij}$ that the vehicle
will meet when it reaches the pair. Two residual message-passing layers~\eqref{eq:mp}, with messages weighted by the learned range kernel~\eqref{eq:kernel}, let
each slot depend on its neighbors; a shared head emits one logit $\ell_i$ per
slot, and the logit signs define the proposed mode $m^{\mathrm{nn}}$.}
\label{fig:avoidnet}
\end{figure}

\section{Imitation Learning of Avoidance Modes}
\label{sec:method}

This section describes how the expert's enumeration is turned into supervision,
how the resulting proposer is structured, and how it is trained. 

\subsection{Expert Protocol and Labels}
\label{sec:labels}

The data are collected by running the expert of Section~\ref{sec:mmmpc} in closed loop with its selection rule~\eqref{eq:hyst}. The label,
however, is the minimum-cost mode $m^{\mathrm{best}}$ rather than the executed
mode, because the network replaces only $m^{\mathrm{best}}$ in the selection rule, which still applies the hysteresis at deployment.
The label assigns one side $\sigma^\star_i$ to every branch slot and comes at no extra cost, because the expert's enumeration already yields $J^*(m)$ for all $K$ modes at every step. Steps at which no mode is accepted
(Section~\ref{sec:mmmpc}) carry no label and are discarded.

Training data consist of $210$ expert rollouts in the collection environment of Section~\ref{sec:setup}, with seeds disjoint from all evaluation seeds, yielding $77{,}925$ labeled decision steps. The split is
made by rollout: $42$ of the $210$ rollouts ($20\%$) are reserved for validation
($14{,}892$ steps). The training portion is doubled to $126{,}066$ steps by a
$y$-reflection augmentation that mirrors each scene about the corridor
centerline and negates every side $\sigma^\star_i$, exploiting the left--right
symmetry of the corridor and of the OCP.

\subsection{Graph Neural Network for Pairwise Clearance}
\label{sec:gnn}

Each side decision concerns one obstacle, so the proposer is built
around a per-slot backbone whose weights are shared across the $N_b$ branch
slots. Slot $i$ carries the $12$-dimensional input
\begin{equation}
\begin{aligned}
\mathbf{z}_i&=\big[\mathbf{z}^{\mathrm{obs}}_i,\;\mathbf{z}^{\mathrm{ego}}\big],\\
\mathbf{z}^{\mathrm{obs}}_i&=\big[p_{i,x}^{B},\;p_{i,y}^{B},\;\rho_i,\;
v_{i,x}^{B},\;v_{i,y}^{B},\;a_i\big],\\
\mathbf{z}^{\mathrm{ego}}&=\big[d_{\mathrm{g}},\;y,\;\sin\psi,\;\cos\psi,\;u,\;r\big],
\end{aligned}
\label{eq:node}
\end{equation}
where the superscript $B$ denotes a quantity rotated by $-\psi$ into the ego body frame. The vector $\mathbf{z}^{\mathrm{obs}}_i$ holds the body-frame relative position and velocity, the radius and the active flag $a_i$ of Section~\ref{sec:mmmpc}, and $\mathbf{z}^{\mathrm{ego}}$, shared across slots, holds the remaining longitudinal distance to the goal
$d_{\mathrm{g}}=x_{\mathrm{goal}}-x$, the lateral position, the heading as a
$\sin/\cos$ pair, the speed and the yaw rate. The network receives these quantities scaled by fixed constants to order
one.

The backbone embeds each slot input $\mathbf{z}_i$ into $\mathbf{h}_i\in\mathbb{R}^d$ with a multilayer perceptron (MLP)
$\mathrm{Lin}(12{\to}d)$--ReLU--$\mathrm{Lin}(d{\to}d)$ and reads one logit
per slot through an MLP head
$\mathrm{Lin}(d{\to}d)$--ReLU--$\mathrm{Lin}(d{\to}1)$. Sharing the weights and reading each
decision at its own slot makes the architecture permutation-equivariant by
construction, so the obstacle-selection rule may re-order slots between steps without
changing the meaning of the output. Without the message-passing layers below, this backbone is already a complete
per-slot proposer; at $d=64$ ($9{,}217$ parameters) it is the Slot-MLP baseline
of Section~\ref{sec:exp}, whereas SideNet uses $d=128$.

The preferred side of obstacle $i$, however, does not depend on that obstacle alone: a neighbor can block the gap that the isolated choice would use, reversing the side while $\mathbf{z}_i$ is unchanged.

\begin{remark}[Structural blindness]
\label{rem:blind}
In a slot-independent architecture, the logit of branch slot $i$ is a function
of $\mathbf{z}_i$ alone. Hence, for any weights, relocating another obstacle
$j$ leaves the logit of obstacle $i$ exactly unchanged.
An edge from $j$ to $i$ carries the missing neighbor dependence directly.
\end{remark}

We therefore make the proposer a GNN: the $N_b$ branch
slots become the nodes of a complete graph with
$\binom{N_b}{2}$ unordered pairs, each used in both message directions, and
edge-conditioned message-passing layers are inserted between the
embedding and the head \citep{gilmer2017neural,battaglia2018relational}.
Fig.~\ref{fig:avoidnet} shows the resulting proposer. The graph spans
the branch slots only: guard obstacles carry no side decision and are not
input to the network, but they remain in both candidate OCPs, so they still
shape the optimized trajectories and the cost-based arbitration.

For moving obstacles, the gap that matters for the side decision is the
one the vehicle meets when it first reaches either obstacle, not the current gap of the pair or the gap at the pair's own closest approach.
Edge $(i,j)$ carries the four features
\begin{equation}
\mathbf{e}_{ij}=\big[g_{ij},\;g^*_{ij},\;\dot d_{ij},\;a_{ij}\big],
\label{eq:edge}
\end{equation}
where $a_{ij}=a_ia_j$ indicates that both slots of the pair are
active. Let
$\mathbf{p}_{ij}=\mathbf{o}_i-\mathbf{o}_j$ and
$\mathbf{v}_{ij}=\mathbf{v}_i-\mathbf{v}_j$. The current and predicted surface
gaps and the center-distance rate are
\begin{equation}
\begin{aligned}
g_{ij}&=\lVert\mathbf{p}_{ij}\rVert-\rho_i-\rho_j,\\
t^*_{ij}&=\min(t_{\mathrm{CPA},i},t_{\mathrm{CPA},j}),\\
g^*_{ij}&=\lVert\mathbf{p}_{ij}+\mathbf{v}_{ij}t^*_{ij}\rVert-\rho_i-\rho_j,\\
\dot d_{ij}&=\frac{\mathbf{p}_{ij}^{\top}\mathbf{v}_{ij}}
{\lVert\mathbf{p}_{ij}\rVert+\epsilon},
\end{aligned}
\label{eq:edge_geometry}
\end{equation}
where $t_{\mathrm{CPA},i}\in[0,T_h]$ is the time of closest approach between the
vehicle and obstacle $i$ under constant velocities, clipped to the horizon, and
$\epsilon>0$ is a small regularizer.  Negative $\dot d_{ij}$
denotes closing. The continuous edge features are scaled in the same
way as the node inputs. All edge features are invariant under the
$y$-reflection augmentation. These are geometric clearance cues, not a
certificate of dynamically feasible passage.

A gap matters only while it is narrow, so each message is weighted by a learned exponential range kernel
\begin{equation}
w_{ij}=\exp\!\big(-\max(g^*_{ij},0)/\kappa\big),\qquad \kappa>0,
\label{eq:kernel}
\end{equation}
where $g^*_{ij}$ is in meters and the length scale $\kappa$ is a trainable parameter, initialized at the bumper-inclusive passage width
$w_s=2(R_s+\delta)=\SI{3.4}{m}$ and kept positive by a softplus
reparameterization. Each message-passing layer then applies the residual
update
\begin{equation}
\mathbf{h}_i\leftarrow \mathbf{h}_i+g_u\Big(\big[\mathbf{h}_i,\;
\textstyle\sum_{j\neq i}a_{ij}\,w_{ij}\,g_m([\mathbf{h}_i,\mathbf{h}_j,
\mathbf{e}_{ij}])\big]\Big),
\label{eq:mp}
\end{equation}
where $g_m$ is a two-layer MLP producing $d_m$-dimensional messages and $g_u$
is a learned affine map. SideNet stacks two such layers at $d=128$ and
$d_m=32$, giving $94{,}851$ parameters. Because the message and update
functions are shared and incoming messages are summed, the GNN preserves the permutation equivariance of the per-slot backbone.

\subsection{\texorpdfstring{Training}{Training}}
\label{sec:train}

SideNet is trained with a binary cross-entropy loss between the logits $\ell_i$ and the targets $(1+\sigma^\star_i)/2$ of Section~\ref{sec:labels}, taken over active branch slots. The loss is minimized with AdamW (learning rate
$10^{-3}$, weight decay $10^{-4}$, cosine-annealed) for $80$ epochs at batch size
$512$ from a fixed seed, and the retained checkpoint is the epoch with the best
validation side accuracy, the fraction of active branch slots whose predicted
side matches the label.

Steps in which a pair is only marginally passable, with a gap between the hull width $2R_s$ and $w_s$, are rare in the collection rollouts. The training set is therefore supplemented with ten short expert episodes, each with a single obstacle pair, at surface gaps from $2.0$ to \SI{4.2}{m}. They add $1{,}888$ steps after reflection ($1.5\%$ of the training set); the validation split is unchanged. All proposers of Section~\ref{sec:exp} receive these episodes; whether one can exploit them depends on its architecture (Section~\ref{sec:case}).

\section{\texorpdfstring{SideNet-MPC: Two-Candidate Deployment}{SideNet-MPC: Two-Candidate Deployment}}
\label{sec:deploy}

SideNet-MPC is the expert of Section~\ref{sec:mmmpc} with a single substitution: the minimum-cost mode $m^{\mathrm{best}}$, which requires all $K$ solves, is replaced by the mode $m^{\mathrm{nn}}$ proposed by the network (Algorithm~\ref{alg:deploy}).  The previous mode is a good candidate for the same reason that underlies real-time iteration (RTI) \citep{diehl2005real} and warm-started mixed-integer MPC \citep{marcucci2020warm}: consecutive OCPs in a receding-horizon loop differ only slightly, so the previous solution usually remains good for the current period. Unlike RTI, which reuses the previous numerical solution, SideNet-MPC reuses the previous mode itself as a candidate and re-solves it for the current obstacle positions.

At each control period, the expert's obstacle-selection rule~\eqref{eq:select} picks the $N_b$ branch and $N_g$ guard obstacles, and a single network forward pass then produces the branch logits, whose signs define the first candidate $m^{\mathrm{nn}}$.

The obstacle-selection rule~\eqref{eq:select} re-ranks the branch slots every period, so a slot may now hold a different obstacle; the second candidate $m^{\mathrm{prev}}$ is therefore rebuilt by obstacle identity rather than copied slot by slot. Each current branch obstacle is matched by world-frame position to the nearest previous branch obstacle within \SI{1.5}{m}; a matched obstacle keeps its previously executed side, and an obstacle that has just entered the branch set takes the side proposed by the network. If no branch obstacle matches, or if the rebuilt $m^{\mathrm{prev}}$ equals $m^{\mathrm{nn}}$, only $m^{\mathrm{nn}}$ is solved.

Both candidates are solved with the expert's OCP~\eqref{eq:ocp}, acados solver and acceptance criterion (Section~\ref{sec:mmmpc}). Warm starts are kept per mode: a mode is initialized from its iterate cached the last time it was accepted, or from the reference trajectory if it has no cache.

When both candidates are accepted, they are arbitrated by the expert's rule~\eqref{eq:hyst} with the same $\eta$: the previous mode is kept unless the network proposal beats it by more than the relative tolerance. If only one candidate is accepted, it is executed; if none is, the input of the candidate solved last is applied, with no fallback or repair step.

\begin{algorithm}[t]
\caption{Two-candidate mode selection for one control period}
\label{alg:deploy}
\begin{algorithmic}[1]
\State $(\mathcal{B},\mathcal{G})\gets$ branch and guard obstacles selected by~\eqref{eq:select}
\State $\boldsymbol{\ell}\gets$ network forward pass on $\mathcal{B}$;
$m^{\mathrm{nn}}\gets\mathrm{sign}(\boldsymbol{\ell})$ on active slots
\State $m^{\mathrm{prev}}\gets m^{\mathrm{nn}}$
\State match each current branch obstacle to the nearest stored one in the world frame ($\le\SI{1.5}{m}$); give each matched obstacle its stored side in
$m^{\mathrm{prev}}$
\State $\mathcal{M}\gets\{m^{\mathrm{nn}}\}$
\If{some obstacle matched \textbf{and} $m^{\mathrm{prev}}\neq m^{\mathrm{nn}}$}
\State $\mathcal{M}\gets\mathcal{M}\cup\{m^{\mathrm{prev}}\}$
\EndIf
\For{$m\in\mathcal{M}$}
\State warm-start from $m$'s cached iterate, or from the reference if no
cache exists; solve~\eqref{eq:ocp}; cache the iterate if accepted
\State $J^*(m)\gets$ cost of the iterate if accepted (SQP success or residuals
$<5\times10^{-2}$), otherwise $\infty$
\EndFor
\State $m^\star\gets m^{\mathrm{prev}}$ if both are accepted and~\eqref{eq:hyst}
holds, else $\argmin_{m\in\mathcal{M}}J^*(m)$
\State apply the input of $m^\star$, or of the candidate solved last if none is accepted
\State store the executed sides and branch-obstacle positions for the next
period
\end{algorithmic}
\end{algorithm}

\section{Simulation Results}
\label{sec:exp}

\subsection{Setup}
\label{sec:setup}

Expert data collection and every closed-loop evaluation use the same planar corridor, with lateral bounds $y\in[-15,15]$~m and the straight reference path $y=0$. The vehicle starts on the reference path, aligned with the positive $x$-axis at the desired speed of \SI{2.5}{m/s}, and a rollout succeeds if it comes within \SI{5}{m} of its goal without collision before the time limit. Ego and obstacle states are observed without noise. Table~\ref{tab:params} lists the controller parameters, which are shared by the expert and by every learned controller.

\begin{figure}[t]
\centering
\begin{subfigure}{\narrowfigw}
\centering
\includegraphics[width=\linewidth]{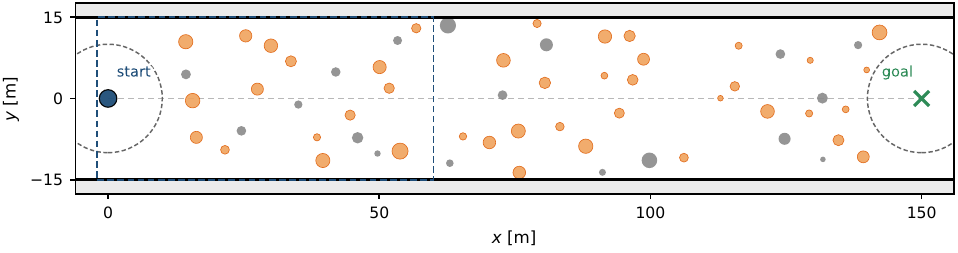}
\caption{Full course with $18$ static (gray) and $42$ dynamic (orange) obstacles. Dashed circles mark the
obstacle-free neighborhoods of the start and goal.}
\end{subfigure}
\vspace{0.45em}
\begin{subfigure}{\narrowfigw}
\centering
\includegraphics[width=0.667\linewidth]{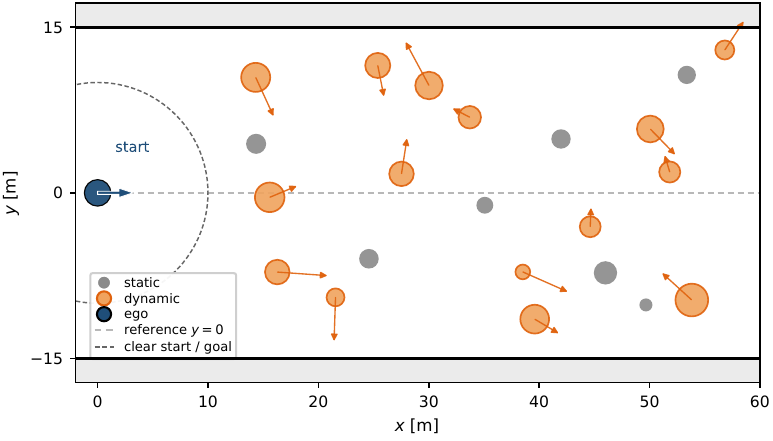}
\caption{Enlargement of the boxed start region. Arrows show the sampled
constant-velocity headings of the dynamic obstacles.}
\end{subfigure}
\caption{Representative initial snapshot of the collection environment.}
\label{fig:simenv}
\end{figure}

\begin{table}[t]
\caption{Simulation and controller parameters.}
\label{tab:params}
\centering
\footnotesize
\setlength{\tabcolsep}{2.6pt}
\renewcommand{\arraystretch}{1.2}
\begin{tabular}{cc|cc}
\toprule
Symbol & Value & Symbol & Value\\
\midrule
$T_s$ & \SI{0.2}{s} & $N$ ($T_h$) & $40$ (\SI{8}{s})\\
$N_b$ / $N_g$ & $4$ / $8$ & $K$ & $16$\\
$\gamma$ & $0.20$ & $\delta$ & \SI{0.5}{m}\\
$R_s$ & \SI{1.2}{m} & $d_{\mathrm{sens}}$ & \SI{20}{m}\\
$c_R$ & $2$ & $r_{\max}$ & \SI{0.80}{rad/s}\\
$a_{\max}$ & \SI{1.0}{m/s^2} & $\dot r_{\max}$ & \SI{1.0}{rad/s^2}\\
$[0,u_{\max}]$ & $[0,3.5]$~m/s & $u_{\mathrm{des}}$ & \SI{2.5}{m/s}\\
$\theta_d$ & \SI{90}{\degree} & $(w_d,w_{\psi})$ & $(0.7,0.3)$\\
$Q$ & $\mathrm{diag}(0.15,0.45,2,3,1)$ & $R_u$ & $\mathrm{diag}(8,8)$\\
$(\lambda_q,\lambda_l)$ & $(5{\times}10^{6},5{\times}10^{3})$ & $\eta$ & $0.20$\\
corridor & $150\times\SI{30}{m}$ & time limit & $80$--\SI{150}{s}\\
obstacle radii & $0.5$--\SI{1.5}{m} & obstacle speeds & $0.5$--\SI{2.0}{m/s}\\
match / goal & \SI{1.5}{m} / \SI{5}{m} & collection & $18$ static, $42$ dyn.\\
\bottomrule
\end{tabular}
\end{table}

The collection environment (Fig.~\ref{fig:simenv}) runs from $(0,0)$ to $(150,0)$ and holds $18$ static and $42$ dynamic circular obstacles, with radii $\mathcal{U}(0.5,1.5)$~m, a minimum surface separation of \SI{2}{m} and an obstacle-free neighborhood of \SI{10}{m} around the start and the goal. Dynamic obstacles receive an isotropic heading $\mathcal{U}(-\pi,\pi)$ and a speed $\mathcal{U}(0.5,2.0)$~m/s and then move at constant velocity, reflecting the normal velocity component at the corridor walls and passing through one another. Traffic is thus unaffected by the controller under test, so the same seed gives every controller identical obstacle trajectories. The environment is deliberately more crowded than any evaluation scene, because its purpose is to visit as many distinct branch configurations per rollout as possible.

\begin{figure}[t]
\centering
\begin{subfigure}{\linewidth}
\centering
\includegraphics[width=\linewidth]{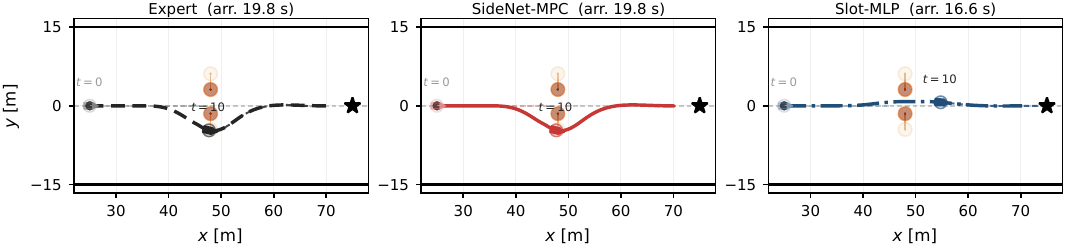}
\caption{Closing gate.}
\label{fig:case-closing}
\end{subfigure}
\vspace{0.3em}
\begin{subfigure}{\linewidth}
\centering
\includegraphics[width=\linewidth]{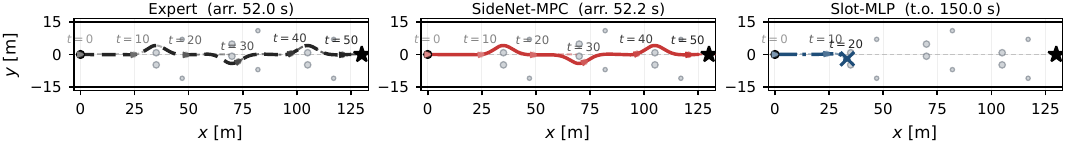}
\caption{Staggered gates.}
\label{fig:case-twin}
\end{subfigure}
\vspace{0.3em}
\begin{subfigure}{\linewidth}
\centering
\includegraphics[width=\linewidth]{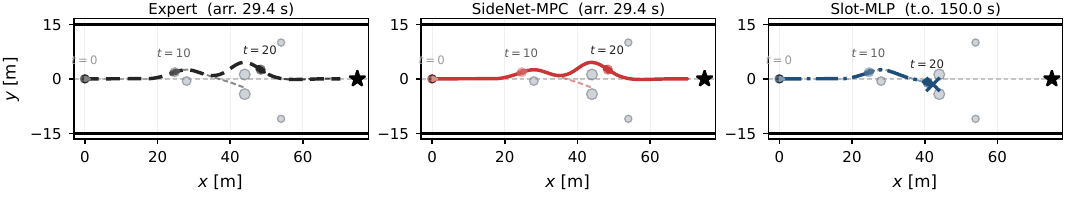}
\caption{Obstacle and gate.}
\label{fig:case-triple}
\end{subfigure}
\caption{Deterministic scenes. Each subfigure shows the expert, SideNet-MPC
and Slot-MLP, with the vehicle and nearby obstacles drawn every \SI{10}{s}; a
cross marks where a run ends without arriving. The expert and SideNet-MPC take
the same topology in all three scenes.}
\label{fig:cases}
\end{figure}

\subsubsection*{\texorpdfstring{Evaluation scenes}{Evaluation scenes}}
Closed-loop evaluation uses three deterministic scenes and one randomized
Monte Carlo scene:
\begingroup\sloppy
\begin{itemize}
\item \textbf{Closing gate}: two obstacles at $x=\SI{48}{m}$ form a gate that
narrows from $8$ to \SI{2}{m}, timed to close when a vehicle at the desired speed
would arrive, so the passable side changes while the vehicle approaches.
\item \textbf{Staggered gates}: three static gates at $x=35$, $70$ and
\SI{105}{m}, offset alternately to either side of the reference path, each
formed by two obstacles whose \SI{2.6}{m} gap is narrower than $w_s$, so that
the vehicle must bypass every gate rather than thread it.
\item \textbf{Obstacle and gate}: a single obstacle just right of the reference
path at $x=\SI{28}{m}$, followed at $x=\SI{44}{m}$ by a gate whose lead obstacle covers the path and leaves a \SI{2.4}{m} gap, equal to the hull width $2R_s$, just to its right; having passed the first obstacle on the left, the vehicle must stay
on that side rather than return to the path through the gap.
\item \textbf{Mixed traffic}: ten obstacles per seed, namely three head-on,
three crossing, two to be overtaken and two static, positioned so that their
encounters concentrate near $x\approx\SI{80}{m}$; one hundred seeds are drawn,
identical across controllers.
\end{itemize}
\endgroup
The closing gate tests a gap that changes over time; in the other two deterministic scenes a gate's gap is too narrow to pass, which a proposer can recognize only by relating the gate's two obstacles.

\subsubsection*{\texorpdfstring{Controllers}{Controllers}}
Five controllers are compared. The four learned controllers share the
training data and recipe of Section~\ref{sec:train} and the deployment of Section~\ref{sec:deploy}, and differ only
in the proposer architecture.
\begingroup\sloppy
\begin{itemize}
\item \textbf{Expert}: the enumeration of Section~\ref{sec:mmmpc}, $K=16$ OCP
solves per control period with the selection rule~\eqref{eq:hyst}.
\item \textbf{SideNet-MPC}: two edge-conditioned message-passing layers with
the learned range kernel~\eqref{eq:kernel}, $d=128$ and $d_m=32$.
\item \textbf{Slot-MLP}: the per-slot backbone of Section~\ref{sec:gnn} alone at $d=64$, which passes no information between obstacles.
\item \textbf{Slot-MLP-wide}: the same backbone widened to $d=214$
($95{,}017$ parameters, matching SideNet), so that a difference from SideNet
cannot be explained by parameter count.
\item \textbf{Deep-set}: symmetric pooling over the slot embeddings
\citep{zaheer2017deep} at $d=64$, which shares
information across obstacles but has no pairwise edges.
\end{itemize}
\endgroup

\begin{figure}[t]
\centering
\includegraphics[width=\linewidth]{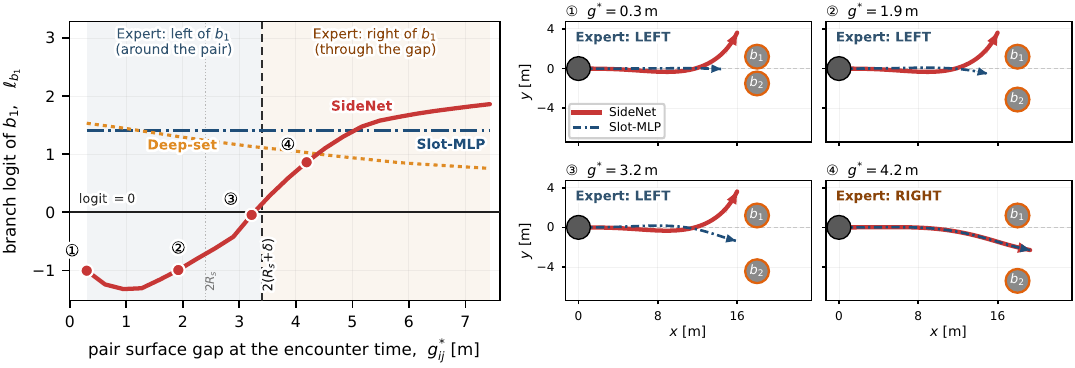}
\caption{Side decision against the pair gap. The partner $b_2$ is slid away from the lead obstacle $b_1$, varying the pair gap $g^{*}$ of~\eqref{eq:edge_geometry} while the ego vehicle and $b_1$ stay fixed.
Left: the branch logit $\ell_{b_1}$, positive for passing $b_1$ on the right.
Shading gives the expert's minimum-cost side; the dashed and dotted vertical lines
mark $w_s=2(R_s+\delta)$ and the hull width $2R_s$. Right: the optimized
trajectories of the modes proposed by SideNet (solid) and Slot-MLP
(dash-dotted) at $g^{*}=0.3$, $1.9$, $3.2$ and \SI{4.2}{m}, the third just
below $w_s$. SideNet's mode is the expert's minimum-cost mode in all four;
Slot-MLP passes $b_1$ on the right in all four, and below $w_s$ its plan heads
into the gap and slows down in front of the pair.}
\label{fig:gapcurve}
\end{figure}

\subsection{\texorpdfstring{Deterministic Scenes}{Deterministic Scenes}}
\label{sec:cases}

\begin{table}[t]
\caption{Deterministic scenes: arrival time [s], where ``timeout'' denotes no
arrival within \SI{150}{s}. The last column is the minimum
clearance on the closing gate.}
\label{tab:cases}
\centering
\footnotesize
\setlength{\tabcolsep}{4.0pt}\renewcommand{\arraystretch}{1.15}
\begin{tabular}{lcccc}
\toprule
& \multicolumn{3}{c}{Arrival time [s]} & Closing gate\\
\cmidrule(lr){2-4}
Controller & Closing & Staggered & Obst.\,\&\,gate & min.\ clear.\ [m]\\
\midrule
\rowcolor{gray!15}
Expert & 19.8 & 52.0 & 29.4 & 0.57\\
\midrule
\textbf{SideNet-MPC} & 19.8 & 52.2 & 29.4 & 0.57\\
Slot-MLP & 16.6 & timeout & timeout & 0.14\\
Slot-MLP-wide & 16.6 & timeout & timeout & 0.14\\
Deep-set & 16.6 & timeout & 29.4 & 0.14\\
\bottomrule
\end{tabular}
\end{table}

Table~\ref{tab:cases} reports the three deterministic scenes and
Fig.~\ref{fig:cases} shows the corresponding trajectories. Only SideNet-MPC and the expert complete all three. The three proposers without pairwise edges time out at the staggered gates, and Slot-MLP and Slot-MLP-wide also time out in the obstacle-and-gate scene.

The closing gate shows the limitation of arrival time alone. The three proposers without pairwise edges arrive first, at \SI{16.6}{s}, but only by passing through the gate before it shuts, with a minimum clearance of \SI{0.14}{m}. SideNet-MPC follows the expert's route south of the gate, with the same arrival time of \SI{19.8}{s} and the same clearance of \SI{0.57}{m}.

The failures in the other two scenes are deadlocks rather than collisions. Fig.~\ref{fig:case-twin}
shows Slot-MLP stopping in front of the first gate, aimed at its blocked gap, and holding position until the \SI{150}{s} timeout; Deep-set stops at the
same point and Slot-MLP-wide in front of the second gate.

In the obstacle-and-gate scene (Fig.~\ref{fig:case-triple}), every controller passes the first obstacle on the left. While the lead obstacle of the gate is the only branch obstacle ahead, the
expert and SideNet-MPC at first plan to pass it on the right, back toward the path;
once its partner becomes a branch obstacle as well, both switch to the left and
bypass the gate. Slot-MLP and Slot-MLP-wide keep the right side, steer into the
gap and stop in front of it at $x\approx\SI{42}{m}$. SideNet-MPC, whose edges
carry the gap between each lead obstacle and its partner, bypasses every gate on
the open side in both scenes.

These deadlocks show what the two-candidate deployment relies on: the previously executed mode
is itself an earlier proposal, so a proposer that keeps choosing the blocked
side supplies both candidates, and the arbitration has no open-side candidate to choose. The proposal must therefore be right, and this is
where the proposer architecture matters.

\begin{figure}[t]
\centering
\includegraphics[width=\linewidth]{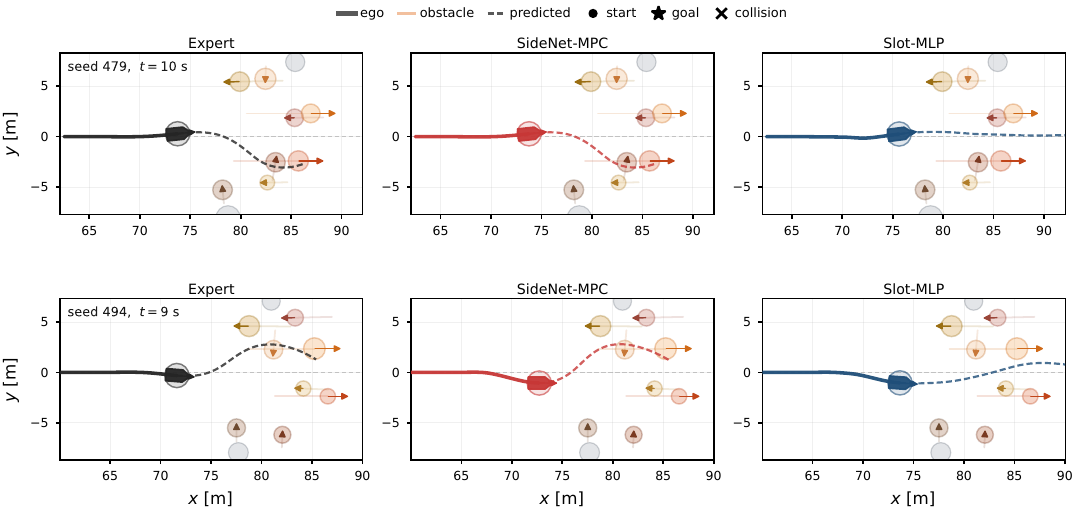}
\caption{Two Monte Carlo seeds (rows) for the expert, SideNet-MPC and Slot-MLP
(columns), each row shown at the instant the encounter is decided. Each panel shows
the past \SI{5}{s} of motion, the obstacles in view with their velocity arrows,
and the predicted path of the mode being executed. The expert and SideNet-MPC
select the same side, while Slot-MLP plans into the gap that a partner obstacle
is closing, north of the expert's route in seed $479$ (top, $t=\SI{10}{s}$) and
south of it in seed $494$ (bottom, $t=\SI{9}{s}$), and collides at
\SI{12.4}{s} and \SI{11.2}{s} respectively.}
\label{fig:mc-grid}
\end{figure}

\subsection{\texorpdfstring{Pairwise Edges and Structural Blindness}{Pairwise Edges and Structural Blindness}}
\label{sec:case}

The deterministic scenes show that the proposers without pairwise edges
fail, but not why. Fig.~\ref{fig:gapcurve} isolates one planning step: with the ego vehicle and the lead branch obstacle $b_1$ fixed, the partner $b_2$ is slid away from $b_1$ so that the pair gap $g^{*}$ sweeps from $0.3$ to \SI{7.4}{m} across $w_s$. The expert, re-solving all sixteen
modes at each sample, switches from the left to the right of $b_1$ between $3.22$ and
\SI{3.54}{m}, an interval that contains $w_s=\SI{3.4}{m}$: it takes the gap once the gap is wide enough for the vehicle plus its bumper margin.

 Slot-MLP and Slot-MLP-wide return $\ell_{b_1}=+1.403$ and $+1.184$, respectively, at every sample: by Remark~\ref{rem:blind}, the partner does not enter $\ell_{b_1}$. Deep-set falls from $+1.533$ to $+0.755$ as the
gap widens and never changes side. SideNet moves from $-1.322$ to $+1.862$ and crosses zero
between $3.22$ and \SI{3.54}{m}, the same interval in which the expert switches.
At the last sample before the switch, $g^{*}=\SI{3.22}{m}$, SideNet still
proposes the expert's mode, whereas Slot-MLP's plan steers into the gap and slows from $2.5$ to \SI{1.2}{m/s} within the horizon (third
trajectory in Fig.~\ref{fig:gapcurve}, right).

The proposers without pairwise edges thus commit to one side of the lead obstacle whatever the gap, which is the failure seen at the staggered gates and, for Slot-MLP and Slot-MLP-wide, in the obstacle-and-gate scene. SideNet, in contrast, has recovered the gap at which the expert switches side and not just its average behavior; since every proposer was trained on the same data, the difference lies in the architecture. Sliding $b_2$ also changes its own node features, but Deep-set
receives these too and never changes side, so the partner's node features alone,
pooled without pairwise edges, do not produce the switch.

\subsection{\texorpdfstring{Monte Carlo Evaluation}{Monte Carlo Evaluation}}
\label{sec:mc}

\begin{table}[t]
\caption{Closed-loop Monte Carlo evaluation on mixed traffic ($100$ seeds,
identical across controllers). Rel.\ SR is the success count divided by the
expert's. Time is the mean computation time per control period
(Section~\ref{sec:comp}), measured with the listed number of CPU threads.}
\label{tab:mix}
\centering
\setlength{\tabcolsep}{4.5pt}\renewcommand{\arraystretch}{1.15}
\begin{tabular}{lcccc}
\toprule
Controller & Succ. & Rel.\ SR & Threads & Time [ms]\\
\midrule
\rowcolor{gray!15}
Expert & 97 & 1.00 & 16 & 35.5\\
\midrule
\textbf{SideNet-MPC} & \textbf{90} & \textbf{0.93} & \textbf{1} & \textbf{18.6}\\
Slot-MLP & 82 & 0.85 & 1 & 19.0\\
Slot-MLP-wide & 83 & 0.86 & 1 & 19.2\\
Deep-set & 80 & 0.82 & 1 & 19.1\\
\bottomrule
\end{tabular}
\end{table}

Table~\ref{tab:mix} reports the randomized benchmark. The expert
completes $97$ of $100$ seeds and SideNet-MPC $90$, or $93\%$ of the expert's
successes. On the $88$ seeds where both succeed, SideNet-MPC arrives \SI{1.73}{s} earlier than the expert on average.

SideNet-MPC completes more runs than every proposer without pairwise edges:
$90$ versus $82$ for Slot-MLP, $83$ for Slot-MLP-wide and $80$ for Deep-set.
These margins amount to a few seeds out of $100$, so the Monte
Carlo benchmark alone does not separate the architectures; the deterministic scenes and the gap sweep above isolate the failure mechanism instead. Neither added capacity nor pooling makes up the difference: Slot-MLP-wide, with a parameter count matched to SideNet's, completes only one run more than Slot-MLP, and Deep-set does no better than Slot-MLP.

Fig.~\ref{fig:mc-grid} shows two seeds with the expert, SideNet-MPC and
Slot-MLP on identical traffic. In each, the vehicle meets the crossing group
near $x\approx\SI{80}{m}$, and what matters is which side of the lead obstacle to take before its partner closes the gap. Slot-MLP picks opposite sides in
the two seeds and is wrong in both. As the gap sweep of Section~\ref{sec:case} shows, the side Slot-MLP picks depends on the lead obstacle alone, so whether that side is blocked is incidental. 

\subsection{Computation}
\label{sec:comp}

Table~\ref{tab:mix} reports the mean computation time per control period, measured for all controllers in one session on an Intel Core Ultra~7 270K Plus (24 cores, no simultaneous multithreading).
SideNet-MPC needs \SI{18.6}{ms} on a single thread, 48\% less than the \SI{35.5}{ms} the expert needs with its sixteen OCPs solved in parallel on sixteen threads. Both fit inside the \SI{200}{ms} control period, but only SideNet-MPC does so without parallel hardware.
The SideNet forward pass and feature construction together take about \SI{0.5}{ms}, so the OCP solves account for $96$ to $98\%$ of the time of every learned controller.

\section{Conclusion}
\label{sec:conclusion}

This paper presented SideNet-MPC, which replaces the exponential mode enumeration of the MM-MPC expert with a GNN proposal while keeping the expert's selection rule, so that at most two TC-CBF OCPs are solved per control period. In $100$ seed-matched Monte Carlo runs of mixed traffic, SideNet-MPC completed $90$ runs against the expert's $97$, while needing \SI{18.6}{ms} per control period on a single thread against the expert's \SI{35.5}{ms} on sixteen parallel threads. Proposers without pairwise edges deadlocked at blocked gates, and in a gap sweep only SideNet switched side in the same gap interval as the expert. Future work will extend the graph to include the guard obstacles and validate the controller on embedded hardware.
\begin{ack}
This work was supported by the Glocal University 30 Project of Kongju National
University in 2026 and by the National Research Foundation of Korea (NRF)
grant funded by the Korea government (MSIT) (No. RS-2026-25490010).
The authors declare no competing interests.
\end{ack}

\bibliographystyle{plainnat}
\bibliography{ref}

\end{document}